\documentclass[lettersize,journal]{IEEEtran}
\usepackage{amsmath,amsfonts}
\usepackage{algorithmic}
\usepackage{algorithm}
\usepackage{array}
\usepackage[caption=false,font=normalsize,labelfont=sf,textfont=sf]{subfig}
\usepackage{textcomp}
\usepackage{stfloats}
\usepackage{url}
\usepackage{verbatim}
\usepackage{graphicx}
\usepackage{cite}
\usepackage{lipsum}
\usepackage{adjustbox}
\usepackage{multirow}
\def\BibTeX{{\rm B\kern-.05em{\sc i\kern-.025em b}\kern-.08em
    T\kern-.1667em\lower.7ex\hbox{E}\kern-.125emX}}

\begin{document}

\title{NP$^2$L: Negative Pseudo Partial Labels Extraction for Graph Neural Networks}

\author{Xinjie Shen,~
  Danyang Wu,~
  Jitao Lu,~
  Junjie Liang,~
  Jin Xu,
  Feiping Nie

  \IEEEcompsocitemizethanks{\IEEEcompsocthanksitem

    Xinjie Shen, Junjie Liang and Jin Xu are with the School of Furture Technology, South China University of Technology, Guangzhou 511442, China (E-mail: frinkleko@gmail.com, 19jjliang22@gmail.com, jinxu@scut.edu.cn)

    Danyang Wu is with the School of Electronic and Information Engineering, Xi'an Jiaotong University, Xi'an 710049, China (E-mail: danyangwu.cs@gmail.com)

    Jitao Lu and Feiping Nie are with the School of Computer Science, the School of Artificial Intelligence, Optics and Electronics (iOPEN), and the Key Laboratory of Intelligent Interaction and Applications (Ministry of Industry and Information Technology), Northwestern Polytechnical University, Xi'an, Shaanxi 710072, China (e-mail: dianlujitao@gmail.com, feipingnie@gmail.com)

  }}

\markboth{Journal of \LaTeX\ Class Files,~Vol.~14, No.~8, August~2021}%
{Shell \MakeLowercase{\textit{et al.}}: A Sample Article Using IEEEtran.cls for IEEE Journals}

% % column for its text to clear the IEEEpubid mark.

\maketitle

\begin{abstract}
  How to utilize the pseudo labels has always been a research hotspot in machine learning. However, most methods use pseudo labels as supervised training, and lack of valid assessing for their accuracy. Moreover, applications of pseudo labels in graph neural networks (GNNs) oversee the difference between graph learning and other machine learning tasks such as message passing mechanism. Aiming to address the first issue, we found through a large number of experiments that the pseudo labels are more accurate if they are selected by not overlapping partial labels and defined as negative node pairs relations. Therefore, considering the extraction based on pseudo and partial labels, negative edges are constructed between two nodes by the negative pseudo partial labels extraction (NP$^2$E) module. With that, a signed graph are built containing highly accurate pseudo labels information from the original graph, which effectively assists GNN in learning at the message-passing level, provide one solution to the second issue. Empirical results about link prediction and node classification tasks on several benchmark datasets demonstrate the effectiveness of our method. State-of-the-art performance is achieved on the both tasks.
\end{abstract}


\section{Introduction}

Pseudo labels are widely used in machine learning as an effective way to utilize data's potential distribution. The main idea is to use the model's prediction results as the genuine labels of the data for supervised training, sharing same loss function and same training procedure. As widely use in semi-supervised\cite{semi-supervised-survey,9889184,10109140} learning or self-supervised\cite{self-sup} learning, pseudo labels are recognized that can be used to improve the performance of the model. However, the pseudo labels are not as accurate as the genuine labels, and the model trained with the pseudo labels is not as good as the model trained with the genuine labels, even causing worse performance than the model without pseudo labels\cite{pseudo-evaluate}. Therefore, it is necessary to find a way to improve the accuracy of the pseudo labels.

\begin{figure}[t]
  \centering

  \includegraphics[width=0.49\textwidth]{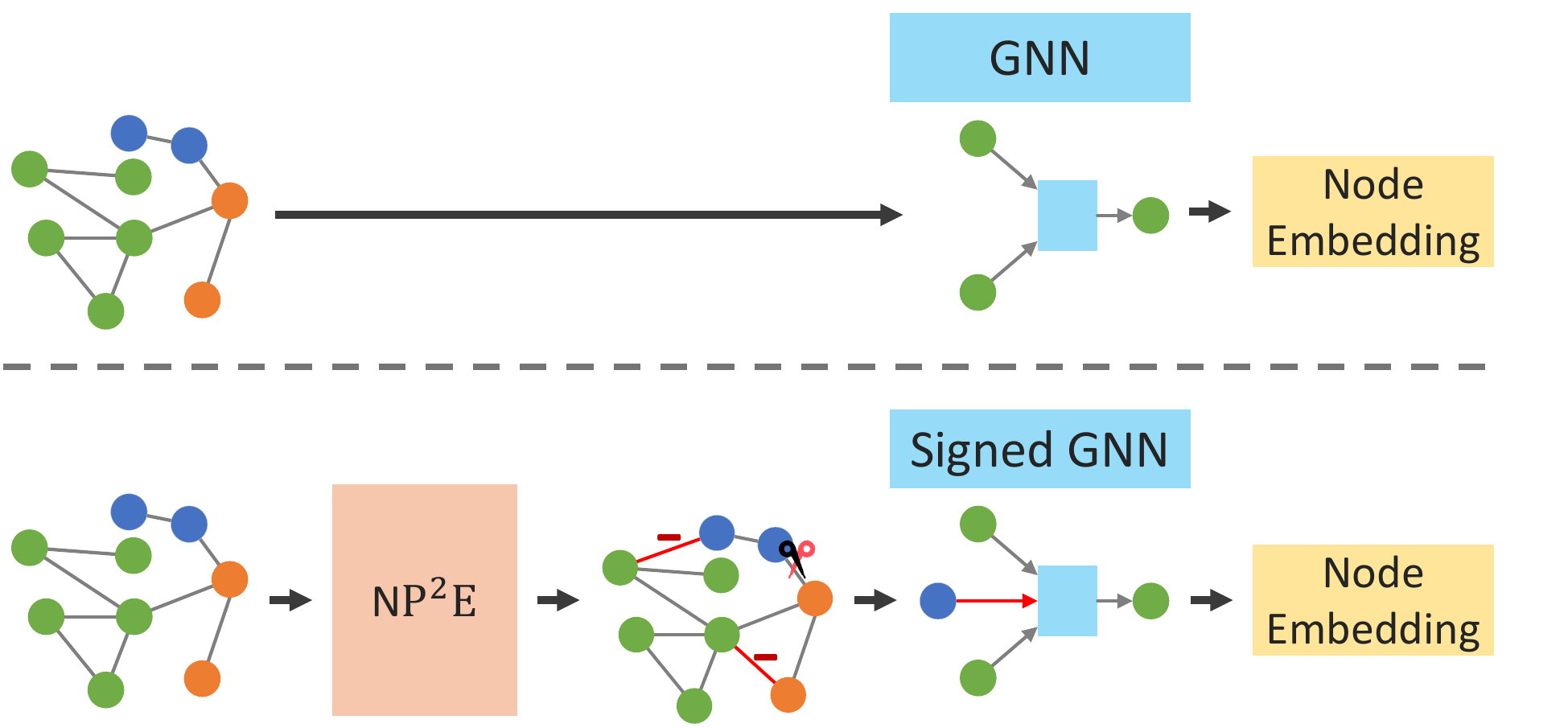}
  \vspace{-0.7cm}
  \caption{The first row shows node embeddings learning process of GNN. The second row shows the process of our method, which use NP$^2$E module to build signed graph and generate node embeddings by learning on the signed graph with Signed GNN.}
  \label{fig:intro}
  \vspace{-0.7cm}
\end{figure}

In the field of GNN, few applications of pseudo labels are proposed in a similar way without considering the significant difference between graph learning and other machine learning tasks such as message passing mechanism. Previous works often pay more attention on designing how to meaningfully generate pseudo labels\cite{supernode,self-enhance,9997579}. Methods like selecting high confidence pseudo labels with threshold\cite{high-threshold} are proposed to improve the qualities of pseudo labels, but suffering from complex tuning of threshold. However, these methods are not effective enough, still suffering from lacking valid assessing for pseudo labels' accuracy and ignoring their distinct contributions comparing with genuine labels to the classification task. Furthermore, the pseudo labels are used without awareness of the graph structure, which is different from the traditional machine learning tasks.

\begin{figure*}[t]
  \centering

  \includegraphics[width=0.99\textwidth]{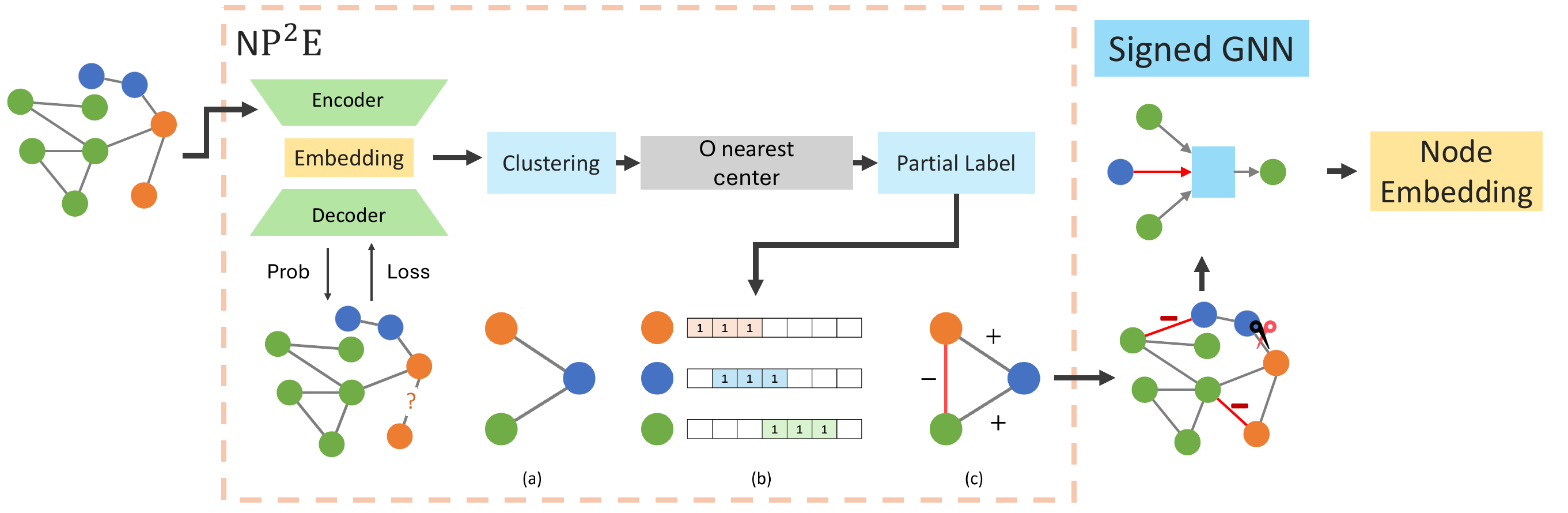}
  \vspace{-0.3cm}
  \caption{The framework of our method. NP$^2$E module which consisted with node embeddings learning and negative pseudo partial labels extraction is applied to extract high quality negative pseudo partial labels and build signed graph. Signed GNN are applied to learn the node embeddings. (a) Original graph (b) Nodes'partial labels (c) Constructed negative edges.}
  \label{fig:framework}
  \vspace{-0.65cm}
\end{figure*}

To this end, we found through a large number of experiments that if the pseudo labels are selected by partial label and defined as negative node pairs, they have better accuracy. Based on this observation, we propose a novel method to extract highly accurate negative pseudo partial labels, building negative edges for GNN's message passing, named as negative pseudo partial labels extraction (NP$^2$E).  As Figure~\ref{fig:framework}, node embeddings are firstly learned by GNN on unsupervised graph reconstructing tasks. Then, one clustering method is applied to generate probability or distance of each node belonging to each cluster. With that, partial labels are defined as nearest top $o$ clusters of each node. With that in mind, negative pseudo partial labels are defined as the negative node pairs relations, which are the node pairs have no overlapping partial labels, as Figure~\ref{fig:framework}b shown. Utilizing the negative nodes pairs relations, negative edges are built based on the original graph, which is shown as Figure~\ref{fig:framework}c. In this way, the original graph is converted into a signed graph, which can be learned by the variants of GNN, like signed GCN\cite{signed-gcn}. Compared with original process of GNN learning, our method can be summarized in Figure~\ref{fig:intro}. The signed graph contains highly accurate pseudo labels information from the original graph, which effectively assists GNN in learning at the message-passing level. Empirical results about link prediction and node classification tasks on several benchmark datasets demonstrate the effectiveness of our method. State-of-the-art performance is achieved on the both tasks.

In summary, our main contributions are summarized as follows:
\begin{enumerate}
  \item A large number of experiments are conducted, proving that negative pseudo partial labels are more accurate.
  \item A simple and effective module NP$^2$E to exact high quality negative pseudo labels and build signed graph are proposed. The module can be easily applied to current GNN pipeline and improve the performance of GNN on message passing level, utilizing negative relations of nodes.
  \item Extensive experiments on several benchmark datasets with signed graph built by NP$^2$E demonstrate the effectiveness of our method. State-of-the-art performance is achieved on the both link prediction and node classification tasks.
\end{enumerate}
\textbf{Notation.} In this paper, we consider two kinds of graph. The first one is the unweighted and undirected graph $\mathcal{G}=(\mathcal{V},\mathcal{E})$, where $\mathcal{V}=\{v_1,v_2,\dots ,v_n \}$ is the set of nodes and $\mathcal{E}\subseteq \mathcal{V} \times \mathcal{V}$ is the set of edges. The second one is the signed graph $\mathcal{G}^s=(\mathcal{V},\mathcal{E}^+, \mathcal{E}^-)$, where $\mathcal{E}^+\subseteq  \mathcal{V} \times \mathcal{V} $ and $\mathcal{E}^-\subseteq  \mathcal{V} \times \mathcal{V} $ are the set of positive edges and negative edges, respectively. Note that $\mathcal{E}^+ \cap \mathcal{E}^- = \emptyset$, which in other words, two nodes would not have both positive and negative edges.
We use $\mathbf{A} \in \mathbb{R}^{n \times n}$ denote the adjacency matrix of $\mathcal{G}$, where $\mathbf{A}_{ij}=1$ if $(v_i,v_j) \in \mathcal{E}$ and $\mathbf{A}_{ij}=0$ otherwise. Similarly, $\mathbf{A} \in \mathbb{R}^{n \times n}$ can also denote the adjacency matrix of a signed graph $\mathcal{G}$, where $\mathbf{A}_{ij}=1$ represent existing one positive link between $v_i$ and $v_j$, $\mathbf{A}_{ij}=-1$ represent existing one negative link between $v_i$ and $v_j$, and $\mathbf{A}_{ij}=0$ represent no link between $v_i$ and $v_j$.

\section{Methodology}
\subsection{Negative Pseudo Partial Labels Extraction}\label{subsec:np2e}
In this section, we introduce the proposed negative pseudo partial labels extraction (NP$^2$E) module. The NP$^2$E module is used to extract negative pseudo partial labels from the original graph $\mathcal{G}$, which can be used to construct signed graph $\mathcal{G}^s$ for GNNs. The pseudo partial labels are defined as negative node pairs relations.

The NP$^2$E module can be summed in Algorithm~\ref{algorithm:BuildSignedGraph}. There are three steps in the NP$^2$E module: 1) Node embedding learning conducted by graph reconstructing task. 2) Pseudo partial labels extraction based on node embedding. 3) Negative pseudo partial labels and signed graph construction.

\subsubsection{Node Embedding Learning}
The first step is to learn node embedding that can represent the original graph $\mathcal{G}$ structure. The goal can be achieved by optimizing graph reconstructing task, which is a typical unsupervised learning task. The goal is to learn node embedding $\mathbf{Z} \in \mathbb{R}^{n \times d}$, where $n$ is the number of nodes and $d$ is the dimension of node embedding, can reconstruct the adjacency matrix $\mathbf{A}$ of the original graph $\mathcal{G}$. It can be formulated as follows:
\begin{equation}
  \begin{aligned}
    \mathbf{Z}             & = f(\mathbf{A},\mathbf{X}), \\
    \mathbf{\widetilde{A}} & = g(\mathbf{Z}),
  \end{aligned}
\end{equation}
where $f$ is the embedding learning function, and $g$ is the graph reconstructing function.

\subsubsection{Pseudo Partial Labels Extraction}
With node embedding, the second step is to extract pseudo partial labels. Firstly, one clustering algorithm is used to cluster nodes embedding $\mathbf{Z}$ into $k$ clusters by minimize the following loss function:
\begin{equation}
  \label{eq:kmeans}
  \mathcal{L}_{cluster} = \sum_{i=1}^{k}\sum_{v_j \in C_i}||\mathbf{Z}_j-\mathbf{C}_i||_2^2.
\end{equation}
Denoted the distance between node embedding $\mathbf{Z}_i$ and cluster center $\mathbf{C}_j$ as $\mathbf{D}_{ij}=||\mathbf{Z}_i-\mathbf{C}_j||_2^2$. Then, each node is assigned to the top $o$ clusters with the smallest distance to generating a partial labels matrix $\mathbf{P} \in \mathbb{R}^{n \times k}$, which can be formulated as follows:
\begin{equation}
  \mathbf{P}_{ij}=\left\{
  \begin{aligned}
    1, & \quad \text{if} \quad \mathbf{D}_{ij} \leq \mathbf{D}_{i(o)} \\
    0, & \quad \text{otherwise},
  \end{aligned}
  \right.
\end{equation}
where $\mathbf{D}_{i(o)}$ is the $o$-th smallest element in $\mathbf{D}_i$.

To this end, we can get an $n \times k$ matrix $\mathbf{P}$, where each row is a zero or one vector, which represents the partial labels of the corresponding node. The number of partial labels is tunable and usually lead to high probability of ground truth fell into the pseudo partial labels when $o$ is large enough. In following section, we will discuss the  reasonable exploitation range of $o$.

\subsubsection{Negative Pseudo Partial labels}
\begin{figure}
  \begin{algorithm}[H]
    \renewcommand{\algorithmicrequire}{\textbf{Input:}}
    \renewcommand{\algorithmicensure}{\textbf{Output:}}
    \caption{Negative Pseudo Partial Labels Extraction}
    \label{algorithm:BuildSignedGraph}
    \begin{algorithmic}[1]
      \REQUIRE Graph $\mathcal{G}=(\mathcal{V},\mathcal{E})$, node feature $\mathbf{X} \in \mathbb{R}^{n \times m}$, number of clusters $k$, partial labels number $o$
      \ENSURE Signed graph $\mathcal{G}^s=(\mathcal{V},\mathcal{E}^+, \mathcal{E}^-)$

      \STATE Train an unsupervised model for link prediction, and get the latent representation $\mathbf{Z} \in \mathbb{R}^{n \times d}$ of nodes,
      $\mathbf{Z} = f(\mathbf{A},\mathbf{X})$.
      % Cluster nodes
      \STATE Cluster nodes embedding $\mathbf{Z}$ into $k$ clusters by K-means  $\mathbf{C} = \arg \min \sum_{i=1}^{k}\sum_{v_j \in C_i}||\mathbf{Z}_j-\mathbf{C}_i||_2^2$.
      % Get distance matrix
      \STATE Get distance matrix $\mathbf{D} \in \mathbb{R}^{n \times k}$,$\mathbf{D}_{ij}=||\mathbf{Z}_i-\mathbf{C}_j||_2^2$.
      % Get partial labels
      \STATE Get partial labels $\mathbf{P} \in \mathbb{R}^{n \times k}$. Assign each node to the top o cluster with the smallest distance. $\mathbf{P}_{ij}=1$ if $\mathbf{D}_{ij} \leq \mathbf{D}_{i(o)}$, $\mathbf{P}_{ij}=0$ otherwise.
      \STATE Get negative pseudo partial labels $\mathbf{N} \in \mathbb{R}^{n \times n}$, where $\mathbf{N}_{ij}=1$ if $v_i$ and $v_j$ don't have any overlapping partial labels. $\mathbf{N}_{ij}=0$ otherwise.
      \STATE Get signed graph $\mathcal{G}^s$, where $\mathcal{E}^+=\{(v_i,v_j)|\mathbf{A}_{ij}=1$ and $\mathbf{N}_{ij}=0\}$ and $\mathcal{E}^-=\{(v_i,v_j)| \mathbf{A}_{ij}=0$ and $\mathbf{N}_{ij}=0\}$.
    \end{algorithmic}
  \end{algorithm}
  \vspace{-0.8cm}
\end{figure}
The third step is to extract negative pseudo partial labels. The pseudo partial labels are defined as negative node pairs relations. Given two nodes $v_i$ and $v_j$, if they don't have any overlapping partial labels, we define them as negative node pairs. Otherwise, we define them as positive node pairs. The negative pseudo partial labels are defined as follows:
\begin{equation}
  \mathbf{N}_{ij}=\left\{
  \begin{aligned}
    % <= oth smallest element in D_i
    1, & \quad \text{if} \quad \mathbf{P}_{i} \mathbf{P}_{j}^T\neq 0 \\
    0, & \quad \text{otherwise}.
  \end{aligned}
  \right.
\end{equation}

As long as the ground truth have quite high probability to fall into the pseudo partial labels, the probability of two nodes with disparate pseudo partial labels to not be the same class is quite high. With increasing the number of partial labels $o$, the probability of ground truth fell into the pseudo partial labels increases and finally reaches 1. It's same for the probability of two nodes with disparate pseudo partial labels to not be the same class.

To use this property into graph learning task, we further propose to utilize the negative pseudo partial labels relations on the original graph $\mathcal{G}$, building negative edges to construct signed graph $\mathcal{G}^s$ for GNNs.

It's natural to use the negative node pairs relations into edge filed. For every two nodes $v_i$ and $v_j$, if $\mathbf{N}_{ij}=1$, we change $\mathbf{A}_{ij} \leftarrow \mathbf{A}_{ij} - 1$. This operation geometrically means that if there is an edge between $v_i$ and $v_j$, we drop it; otherwise, we add a negative edge between $v_i$ and $v_j$. In this way, we can construct signed graph $\mathcal{G}^s$ for GNNs.

\subsection{NP$^2$E-Simple}
In this section, we introduce a simple implementation of NP$^2$E module, which is called NP$^2$E-Simple.

For node embedding learning, we use GAE and VGAE with GCN as encoder. GAE (Graph Auto-Encoder) is an often used GNN for generating node embedding, aiming to reconstruct the adjacency matrix $\mathbf{A}$ of the original graph $\mathcal{G}$. GAE can be formulated as follows:
\begin{equation}
  \begin{aligned}
    \mathbf{Z}             & = Encoder(\mathbf{A},\mathbf{X}),                     \\
    \mathbf{\widetilde{A}} & = Decoder(\mathbf{Z})=\sigma(\mathbf{Z}\mathbf{Z}^T),
  \end{aligned}
\end{equation}
where Encoder are usually consisted of several layers of GCN, and Decoder is a simple inner product decoder.
Training GAE is to minimize the following loss function:
\begin{equation}
  \mathcal{L}_{GAE} = \frac{1}{n^2}\sum_{i,j=1}^{n}(BCE(\mathbf{A}_{ij},\mathbf{\widetilde{A}}_{ij})),
\end{equation}
where $BCE$ is the binary cross entropy loss function.

For VGAE (Variational Graph Auto-Encoder), it use variational inference to learn the distribution of node embedding $\mathbf{Z}$, which can be formulated as follows:
\begin{equation}
  \begin{aligned}
    \mathbf{\mu},\mathbf{\sigma} & = Encoder(\mathbf{A},\mathbf{X}),                                                                              \\
    \mathbf{Z}                   & = \mathbf{\mu}+\mathbf{\sigma}\odot \mathbf{\epsilon}, \quad \mathbf{\epsilon} \sim \mathcal{N}(0,\mathbf{I}), \\
    \mathbf{\widetilde{A}}       & = Decoder(\mathbf{Z})=\sigma(\mathbf{Z}\mathbf{Z}^T).
  \end{aligned}
\end{equation}
Training VGAE is to minimize the following loss function:
\begin{equation}
  \begin{aligned}
    \mathcal{L}_{VGAE} & = \frac{1}{n^2}\sum_{i,j=1}^{n}(BCE(\mathbf{A}_{ij},\mathbf{\widetilde{A}}_{ij}))             \\
                       & +\frac{1}{2}\sum_{i=1}^{n}(1+\log(\mathbf{\sigma}_i^2)-\mathbf{\mu}_i^2-\mathbf{\sigma}_i^2).
  \end{aligned}
\end{equation}

As Equation~\ref{eq:kmeans}, nodes embedding $\mathbf{Z}$ are clustered into $k$ clusters. Then, each node is assigned to the top $o$ clusters with the smallest distance to generating a partial labels matrix $\mathbf{P}$. Finally, we get negative pseudo partial labels matrix $\mathbf{N}$ by $\mathbf{N}_{ij}=1$ if $v_i$ and $v_j$ don't have any overlapping partial labels, $\mathbf{N}_{ij}=0$ otherwise.

\subsubsection{Accuracy of Pseudo Partial Labels}

\begin{figure}[t]
  \centering
  \hspace{-0.4cm}
  \includegraphics[width=0.48\textwidth]{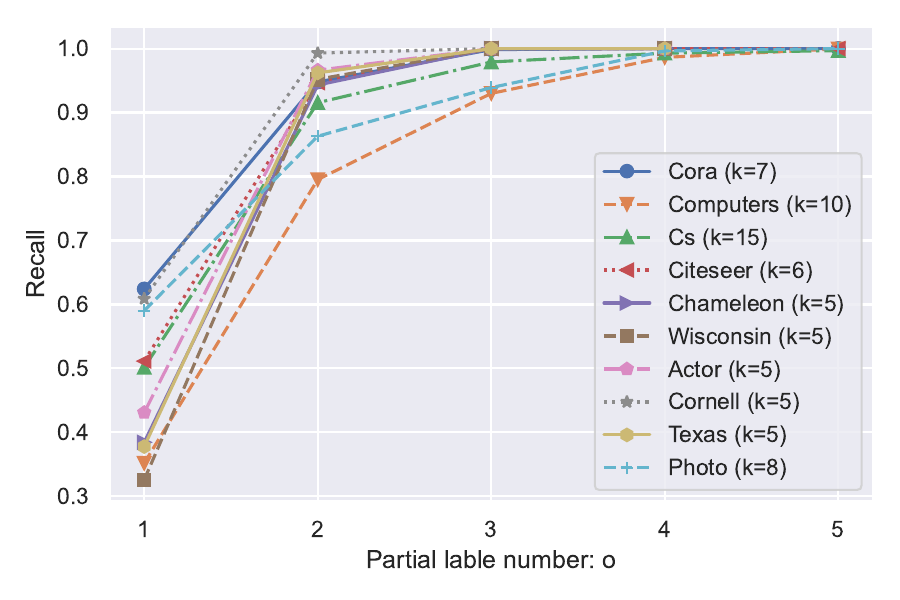}
  \vspace{-0.5cm}
  \caption{Recall score of pseudo partial labels with different $o$ on different datasets.}
  \label{fig:recall}
  \vspace{-0.6cm}
\end{figure}

To explain the pseudo partial labels extraction process and the functionability of the NP$^2$E module, we'd like to introduce the accuracy of pseudo partial labels first. As shown in Algorithm~\ref{algorithm:BuildSignedGraph}, the pseudo partial labels $\mathbf{P}$ is an $n \times k$ matrix, where $n$ is the number of nodes and $k$ is the number of clusters. Each row of $\mathbf{P}$ is a zero or one vector, which represents the partial labels of the corresponding node.

For clear discussion of the accuracy of pseudo partial labels, we need to define one metric which can reflect the accuracy of ground truth fell into the pseudo partial labels. In this paper, we use recall for instance. Let $\mathbf{Y} \in \mathbb{R}^{n \times k}$ denote the ground truth of partial labels, where $\mathbf{Y}_{ij}=1$ if $v_i$ belongs to the $j$-th class, and $\mathbf{Y}_{ij}=0$ otherwise. So that, a recall score can be defined as follows:
\begin{equation}
  recall = \sum_{i=1}^n\sum_{j=1}^n \frac{sign(m_{ij}·s_{ij})}{||\mathbf{S}||_0},
\end{equation}
where $\mathbf{M} = \mathbf{Y}  \mathbf{Y}^T= \{m_{ij}\}, \mathbf{S} = \mathbf{P}  \mathbf{P}^T= \{s_{ij}\}$. $sign(x)$ is the sign function, which is defined as follows:
\begin{equation}
  sign(x)=\left\{
  \begin{aligned}
    % <= oth smallest element in D_i
    1, & \quad \text{if} \quad x > 0 \\
    0, & \quad \text{otherwise}.
  \end{aligned}
  \right.
\end{equation}

After a large number of experiments, we found that the recall score of pseudo partial labels is related to the number of clusters $k$ and the number of partial labels $o$. As shown in Figure~\ref{fig:recall}, the recall score of pseudo partial labels increases with the increase of $o$, achieving quite high recall score when $o$ is large enough, usually between $\frac{k}{3}$ and $\frac{k}{2}$. In following section, we will tune $o$ in this range. Therefore, we can conclude that the pseudo partial labels and negative pseudo partial labels extracted by NP$^2$E module are quite accurate, for the ground truth have quite high probability to fall into the pseudo partial labels.

Two obvious advantages of newly built signed graph to the original graph are as following:
\begin{enumerate}
  \item The newly built signed graph based on more accurate pseudo partial labels information.
  \item Negative edge can bridge different split sets in downstream tasks, which is beneficial to the generalization ability of the model, especially in transductive learning.
\end{enumerate}

\subsection{Utilizing Negative Edges in Message Passing Based GNNs}
\label{subsec:gnn}

Comparing to common GNN used for original unsigned graph, new built signed graph need to be processed by GNNs with negative edges.

Moreover, inspired by ``two stream idea", we proposed a simple architecture that can be used, which can be formulated as follows:
\begin{equation}
  \begin{aligned}
    \mathbf{Z}^{k} & =(1- a^{k}) \cdot f^{k}(\mathbf{X},\mathbf{A}^{pos}) + a^{k} \cdot f^{k}(\mathbf{X},\mathbf{A}^{neg}), \\
    a^{k}          & = \sigma (W_{att}^{k} \cdot f^{k}(\mathbf{X},\mathbf{A}^{neg}).
  \end{aligned}
\end{equation}
where $f$ can be any k-layer GNNs. $\mathbf{A}^{neg}_{ij} = 1$ if $\mathbf{A}_{ij} = -1$, otherwise $\mathbf{A}^{neg}_{ij} = 0$. $\mathbf{A}^{pos}$ is the adjacency matrix of positive edges. $a_i$ is the simple weight balancing positive and negative streams. One example of $f$ can be GNCN \cite{VGNAE}, consisted with one normalizing flow layer and one APPNP layer. Layer of GNCN can be formulated as follows:
\begin{equation}
  \begin{split}
    \begin{aligned}
      \mathbf{L}_v       & = s \frac{\mathbf{Z^{k}_v}}{||\mathbf{Z^{k}_v}||},                                                                    \\
      \mathbf{Z}^{k+1}_v & = \frac{1}{d_v + 1} \mathbf{l}_v  +  \sum_{u \in \mathcal{N}(v)}\frac{1}{\sqrt{d_v + 1}\sqrt{d_u + 1}}  \mathbf{L}_u,
    \end{aligned}
  \end{split}
\end{equation}
where s is a tunable scale factor.

On the other hand, we are lucky that there are some GNNs that can be directly used for signed graph, such as SGCN\cite{signed-gcn}. SGCN (Signed Graph Neural Network) is a typical GNN for signed graph, which is introduced in Appendix~\ref{appendix:sgcn}.

With these GNNs, we can easily perform tasks on signed graph. Link prediction is a typical unsupervised learning task, where can be formulated as follows:
\begin{equation}
  \begin{split}
    \begin{aligned}
      \mathbf{Z}             & = f(\mathbf{A}^s,\mathbf{X}),                                                   \\
      \mathbf{\widetilde{A}} & = \sigma(\mathbf{Z}\mathbf{Z}^T)                                                \\
      \mathcal{L}_{Link}     & = \frac{1}{n^2}\sum_{i,j=1}^{n}(\mathbf{A}_{ij}-\mathbf{\widetilde{A}}_{ij})^2.
    \end{aligned}
  \end{split}
\end{equation}
% supervised task node classification
Node classification is a typical supervised learning task, where can be formulated as follows:
\begin{equation}
  \begin{split}
    \begin{aligned}
      \mathbf{Z}                   & = f(\mathbf{A}^s,\mathbf{X})          \\
      \mathcal{L}_{Classification} & = CrossEntropy(\mathbf{Z},\mathbf{Y})
    \end{aligned}
  \end{split}
\end{equation}

\subsection{Discussion}
Note that the NP$^2$E module can be easily implemented by current methods,  aforementioned K-means and GAE are one simple and typical example. To extend it, we can use more complex methods to extract pseudo partial labels. For node embedding, an alternative method based on contrastive learning is using two unshared weight GNN/MLP to generate two versions of node embedding and reconstructing graph with their inner product, finally take the average of them as the final node embedding. Moreover, multi numbers of clustering method like spectral clustering and hierarchical clustering can be applied and generate consensus pseudo partial labels by voting or other ensemble methods.

\section{Experiments}

\subsection{Benchmark Datasets}

\begin{table}[t]
  \centering
  \vspace{-0.2cm}
  \caption{Statistics of benchmark datasets}
  \vspace{-0.2cm}
  \begin{adjustbox}{width=\columnwidth,center}
    \begin{tabular}{|c|c|c|c|c|c|}
      \hline
      \textbf{Dataset} & \textbf{Type}                 & \textbf{Nodes} & \textbf{Edges} & \textbf{Features} & \textbf{Classes} \\ \hline
      Photo            & \multirow{6}{*}{Homophilic}   & 7,650          & 238,162        & 745               & 8                \\ \cline{1-1} \cline{3-6}
      Computers        &                               & 13,752         & 491,722        & 767               & 10               \\ \cline{1-1} \cline{3-6}
      CS               &                               & 18,333         & 163,788        & 6,805             & 10               \\ \cline{1-1} \cline{3-6}
      Actor            &                               & 7,600          & 30,019         & 932               & 5                \\ \cline{1-1} \cline{3-6}
      Cora             &                               & 2,708          & 10,556         & 1,433             & 7                \\ \cline{1-1} \cline{3-6}
      CiteSeer         &                               & 3,327          & 9,104          & 3,703             & 6                \\ \hline
      Chameleon        & \multirow{5}{*}{Heterophilic} & 2,277          & 36,101         & 2,325             & 5                \\ \cline{1-1} \cline{3-6}
      Squirrel         &                               & 5,201          & 217,073        & 2,089             & 5                \\ \cline{1-1} \cline{3-6}
      Cornell          &                               & 183            & 298            & 1,703             & 5                \\ \cline{1-1} \cline{3-6}
      Texas            &                               & 183            & 325            & 1,703             & 5                \\ \cline{1-1} \cline{3-6}
      Wisconsin        &                               & 251            & 515            & 1,703             & 5                \\ \hline
    \end{tabular}
  \end{adjustbox}
  \label{tab:dataset statistics}
  \vspace{-0.4cm}
\end{table}

In this section, we discuss several datasets included in pytorch-geometric \cite{pyg}. Their statistics information are shown in Table~\ref{tab:dataset statistics}. In Appendix B, we explain the details of these datasets.

\subsection{Ablation Study of Link Prediction}

\begin{table}[t]
  \centering
  \caption{GCN Link prediction results on datasets}
  \vspace{-0.2cm}
  %\multirow{2}{*}{\begin{tabular}[c]{@{}c@{}}Amazon \\ Photo\end{tabular}}
  \begin{adjustbox}{width=\columnwidth,center}
    \begin{tabular}{|c|c|c|c|c|c|}
      \hline
      \multirow{2}{*}{\textbf{Dataset}}                                            & \multirow{2}{*}{\textbf{Metrics}} & \multicolumn{2}{c|}{\textbf{GAE}} & \multicolumn{2}{c|}{\textbf{VGAE}}                                                      \\ \cline{3-6}
                                                                                   &                                   & \multicolumn{1}{l|}{\textbf{GCN}} & \textbf{SGCN}                      & \multicolumn{1}{l|}{\textbf{GCN}} & \textbf{SGCN}  \\ \hline
                                                                                   \multirow{2}{*}{Amazon Photo}    & AUC                               & 94.38                             & \textbf{95.67}                     & 88.43                             & 95.30          \\ \cline{2-6}
                                                                                   & AP                                & 93.69                             & \textbf{94.89}                     & 88.09                             & 94.54          \\ \hline
                                                                                  \multirow{2}{*}{Amazon Computer} & AUC                               & 92.25                             & \textbf{93.86}                     & 91.24                             & 93.51          \\ \cline{2-6}
                                                                                   & AP                                & 92.22                             & \textbf{93.31}                     & 91.49                             & 92.97          \\ \hline
      \multirow{2}{*}{Coauthor CS}    & AUC                               & 93.10                             & 94.58                              & 93.13                             & \textbf{94.80} \\ \cline{2-6}
                                                                                   & AP                                & 92.37                             & 93.83                              & 92.39                             & \textbf{94.03} \\ \hline
      \multirow{2}{*}{Cora}                                                        & AUC                               & 91.00                             & \textbf{91.22}                     & 89.90                             & 90.67          \\ \cline{2-6}
                                                                                   & AP                                & 90.64                             & \textbf{92.04}                     & 90.91                             & 91.55          \\ \hline
      \multirow{2}{*}{CiteSeer}                                                    & AUC                               & 85.92                             & 87.30                              & 84.89                             & \textbf{88.29} \\ \cline{2-6}
                                                                                   & AP                                & 86.57                             & 88.52                              & 85.19                             & \textbf{88.90} \\ \hline
      \multirow{2}{*}{Chameleon}                                                   & AUC                               & 97.48                             & \textbf{98.16}                     & 96.71                             & 97.91          \\ \cline{2-6}
                                                                                   & AP                                & 97.45                             & \textbf{98.13}                     & 96.86                             & 97.91          \\ \hline
      \multirow{2}{*}{Squirrel}                                                    & AUC                               & 93.91                             & \textbf{95.15}                     & 93.71                             & 94.52          \\ \cline{2-6}
                                                                                   & AP                                & 94.95                             & \textbf{95.91}                     & 94.81                             & 95.48          \\ \hline
    \end{tabular}
  \end{adjustbox}
  \label{tab:GCN-link-prediction-results}
  \vspace{-0.5cm}
\end{table}

\begin{table}[t]
  \centering
  \caption{GNCN Link prediction results on datasets. $*$ means the result is not available for GPU (V100 16G) memory issue}
  \vspace{-0.2cm}
  \begin{adjustbox}{width=\columnwidth,center}
    \begin{tabular}{|c|c|c|c|c|c|}
      \hline
      \multirow{2}{*}{\textbf{Dataset}}                                            & \multirow{2}{*}{\textbf{Metrics}} & \multicolumn{2}{c|}{\textbf{GAE}}  & \multicolumn{2}{c|}{\textbf{VGAE}}                                                       \\ \cline{3-6}
                                                                                   &                                   & \multicolumn{1}{l|}{\textbf{GNCN}} & \textbf{SGNCN}                     & \multicolumn{1}{l|}{\textbf{GNCN}} & \textbf{SGNCN} \\ \hline
      \multirow{2}{*}{\begin{tabular}[c]{@{}c@{}}Amazon \\ Photo\end{tabular}}     & AUC                               & 96.77                              & \textbf{96.79}                     & 96.29                              & 96.31          \\ \cline{2-6}
                                                                                   & AP                                & 96.13                              & \textbf{96.15}                     & 95.49                              & 95.49          \\ \hline
      \multirow{2}{*}{\begin{tabular}[c]{@{}c@{}}Amazon \\ Computers\end{tabular}} & AUC                               & 95.83                              & \textbf{95.86}                     & 80.69                              & 80.69          \\ \cline{2-6}
                                                                                   & AP                                & 95.42                              & \textbf{95.47}                     & 81.01                              & 81.01          \\ \hline
      \multirow{2}{*}{\begin{tabular}[c]{@{}c@{}}Coauthor \\ CS\end{tabular}}      & AUC                               & \textbf{96.41}                     & 96.39                              & 94.29                              & $*$            \\ \cline{2-6}
                                                                                   & AP                                & \textbf{95.91}                     & 95.89                              & 93.27                              & $*$            \\ \hline
      \multirow{2}{*}{Cora}                                                        & AUC                               & 94.90                              & \textbf{95.18}                     & 94.25                              & 94.22          \\ \cline{2-6}
                                                                                   & AP                                & 95.45                              & \textbf{95.57}                     & 93.90                              & 93.76          \\ \hline
      \multirow{2}{*}{CiteSeer}                                                    & AUC                               & 96.97                              & \textbf{97.17}                     & 96.03                              & 95.48          \\ \cline{2-6}
                                                                                   & AP                                & 97.21                              & \textbf{97.35}                     & 95.76                              & 95.53          \\ \hline
      \multirow{2}{*}{Chameleon}                                                   & AUC                               & \textbf{98.45}                     & 98.43                              & 97.90                              & 98.20          \\ \cline{2-6}
                                                                                   & AP                                & \textbf{98.18}                     & 98.17                              & 97.41                              & 97.89          \\ \hline
      \multirow{2}{*}{Squirrel}                                                    & AUC                               & 96.65                              & \textbf{96.68}                     & 88.24                              & 88.24          \\ \cline{2-6}
                                                                                   & AP                                & 96.60                              & \textbf{96.64}                     & 85.07                              & 85.07          \\ \hline
    \end{tabular}
  \end{adjustbox}

  \label{tab:GNAE-link-prediction-results}
  \vspace{-0.6cm}
\end{table}

\subsubsection{Models and Settings}
For link prediction task, we compare our method and GCN as encoder in both GAE and VGAE. Moreover, as mentioned as above, we also compare a signed version of state-of-the-art architecture GNCN with itself, using the ``two stream" idea, denoted as SGNCN.

Following \cite{VGNAE}'s setting, edges are split into $\alpha ,\frac{2(1-\alpha)}{3},\frac{1-\alpha}{3}$ for training, validation, and testing. We set $\alpha$ to 0.8. AUC and AP are used as metrics. For each model, we set 2 layers and 128 hidden units. We use Adam optimizer\cite{adam} and search hyperparameter grid $\{0.001, 0.01, 0.05, 0.1, 0.5\}$ for learning rate and $\{0, $1e-4$, $1e-3$, $1e-2$, $5e-2$, $1e-1$, $5e-1$, 1\}$ for weight decay.

\subsubsection{Results for Link Prediction}
Results are shown in Table~\ref{tab:GCN-link-prediction-results} and Table~\ref{tab:GNAE-link-prediction-results}. Considering GNNs of both GCN and state-of-the-art architecture GNCN. In general, our method achieves the best performance in most datasets, significantly helping the baseline models achieve better performance. It demonstrates that our method is effective in link prediction task and applicability to different GNN architectures. Apart from CS and Chameleon dataset, we obtain better results by extending state-of-the-art architecture GNCN into signed graph filed.
\begin{figure*}[h]
  \centering
  \vspace{-0.3cm}
  \captionsetup[subfigure]{font=scriptsize}
  \subfloat[Wrong negative edges ratios of homophilic datasets.]{\includegraphics[width = 0.24\textwidth]{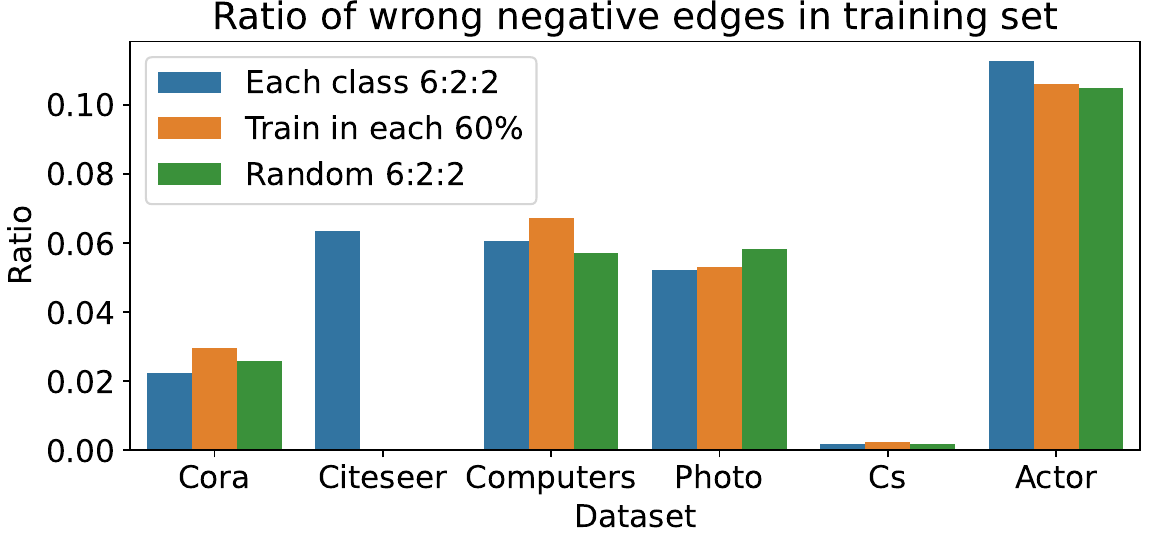}    \label{fig:WrongInTrain}}
  \hfill
  \subfloat[Wrong negative edges ratios of heterophilic datasets.]{\includegraphics[width = 0.24\textwidth]{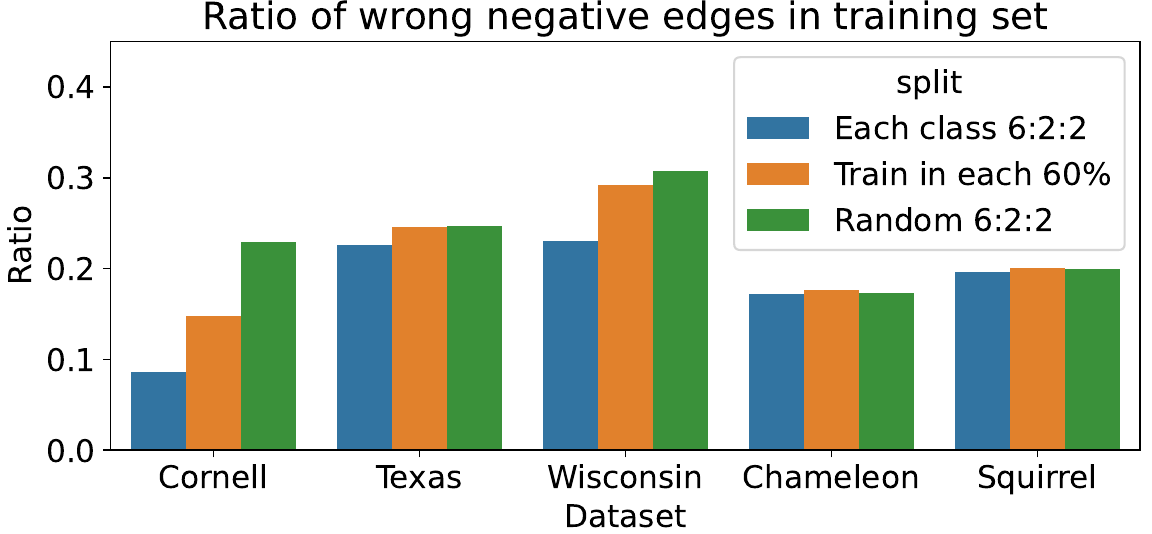}    \label{fig:WrongInTrainHete}}
  \hfill
  \subfloat[Bridging negative edges ratio of homophilic datasets.]{\includegraphics[width = 0.24\textwidth]{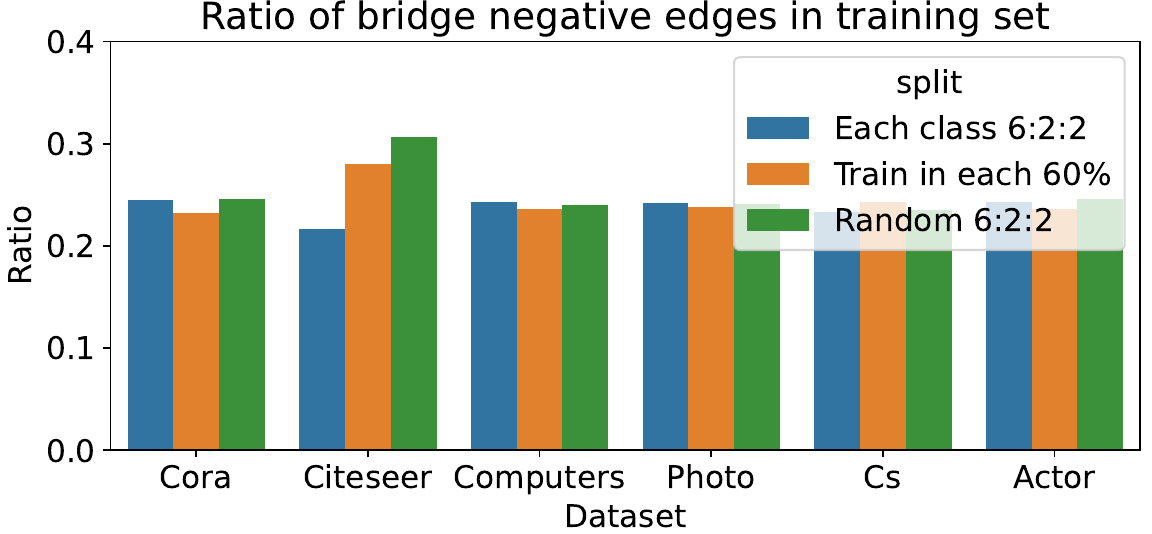}    \label{fig:bridge}}
  \hfill
  \subfloat[Bridging negative edges ratio of heterophilic datasets.]{\includegraphics[width = 0.24\textwidth]{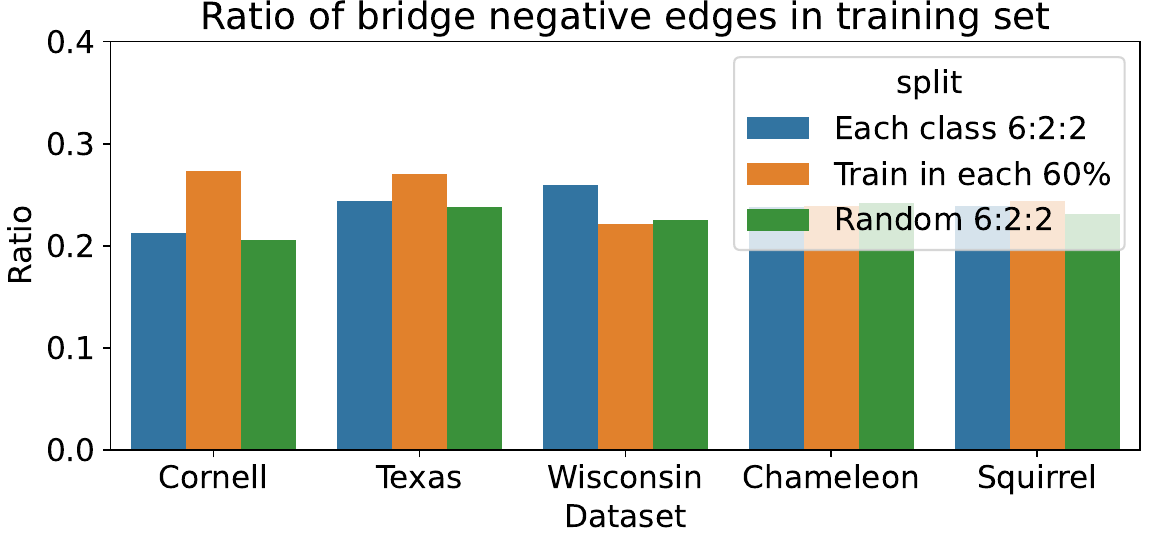}    \label{fig:bridgeHete}}
  \caption{Wrong negative edges ratios and bridging negative edges ratios of different datasets.}
  \vspace{-0.68cm}
\end{figure*}
\subsection{Ablation Study of Node Classification}
\begin{table}[t]
  \centering
  \caption{Node classification results on datasets}
  \vspace{-0.2cm}
  \begin{adjustbox}{width=\columnwidth,center}
    \begin{tabular}{|c|c|c|c|c|}
      \hline
      \textbf{Dataset}                                                           & \textbf{Spilt} & \textbf{GCN}     & \textbf{SGCN}             & \textbf{SGCN$+$}          \\ \hline
      \multirow{3}{*}{\begin{tabular}[c]{@{}c@{}}Amazon\\ Photo\end{tabular}}    & 1              & 93.72 $\pm$ 0.99 & 95.48 $\pm$ 0.43          & \textbf{95.59 $\pm$ 0.36} \\ \cline{2-5}
                                                                                 & 2              & 93.25 $\pm$ 0.32 & 95.09 $\pm$ 0.62          & \textbf{95.46 $\pm$ 0.41} \\ \cline{2-5}
                                                                                 & 3              & 93.88 $\pm$ 0.44 & 95.66 $\pm$ 0.34          & \textbf{95.76 $\pm$ 0.41} \\ \hline
      \multirow{3}{*}{\begin{tabular}[c]{@{}c@{}}Amazon\\ Computer\end{tabular}} & 1              & 90.34 $\pm$ 0.48 & \textbf{90.88 $\pm$ 0.49} & 90.83 $\pm$ 0.51          \\ \cline{2-5}
                                                                                 & 2              & 87.00 $\pm$ 0.55 & 89.77 $\pm$ 0.64          & \textbf{91.07 $\pm$ 0.52} \\ \cline{2-5}
                                                                                 & 3              & 90.35 $\pm$ 0.36 & 90.82 $\pm$ 0.29          & \textbf{90.83 $\pm$ 0.31} \\ \hline
      \multirow{3}{*}{\begin{tabular}[c]{@{}c@{}}Coauthor\\ CS\end{tabular}}     & 1              & 90.83 $\pm$ 0.19 & 94.54 $\pm$ 0.25          & \textbf{94.89 $\pm$ 0.44} \\ \cline{2-5}
                                                                                 & 2              & 94.57 $\pm$ 0.25 & \textbf{95.35 $\pm$ 0.25} & 94.87 $\pm$ 0.31          \\ \cline{2-5}
                                                                                 & 3              & 93.45 $\pm$ 0.31 & 94.81 $\pm$ 0.51          & \textbf{94.97 $\pm$ 0.45} \\ \hline
      \multirow{3}{*}{Actor}                                                     & 1              & 29.91 $\pm$ 0.45 & 36.21 $\pm$ 0.99          & \textbf{36.32 $\pm$ 0.90} \\ \cline{2-5}
                                                                                 & 2              & 29.41 $\pm$ 0.89 & 36.17 $\pm$ 0.82          & \textbf{36.31 $\pm$ 0.72} \\ \cline{2-5}
                                                                                 & 3              & 29.80 $\pm$ 1.17 & \textbf{36.47 $\pm$ 0.93} & 36.32 $\pm$ 1.02          \\ \hline
      \multirow{3}{*}{Chameleon}                                                 & 1              & 49.77 $\pm$ 1.77 & \textbf{52.87 $\pm$ 1.49} & 52.66 $\pm$ 1.68          \\ \cline{2-5}
                                                                                 & 2              & 40.11 $\pm$ 2.07 & \textbf{52.27 $\pm$ 1.97} & 51.40 $\pm$ 1.66          \\ \cline{2-5}
                                                                                 & 3              & 40.24 $\pm$ 2.43 & \textbf{52.53 $\pm$ 1.67} & 52.35 $\pm$ 1.47          \\ \hline
      \multirow{3}{*}{Cornell}                                                   & 1              & 49.49 $\pm$ 4.37 & \textbf{75.38 $\pm$ 5.57} & 74.10 $\pm$ 4.43          \\ \cline{2-5}
                                                                                 & 2              & 51.28 $\pm$ 4.68 & 76.15 $\pm$ 5.80          & \textbf{76.67 $\pm$ 5.60} \\ \cline{2-5}
                                                                                 & 3              & 50.00 $\pm$ 5.24 & \textbf{78.89 $\pm$ 6.03} & 77.84 $\pm$ 6.72          \\ \hline
      \multirow{3}{*}{Texas}                                                     & 1              & 57.95 $\pm$ 3.86 & \textbf{80.51 $\pm$ 5.30} & 80.26 $\pm$ 3.83          \\ \cline{2-5}
                                                                                 & 2              & 59.75 $\pm$ 4.63 & 82.75 $\pm$ 4.16          & \textbf{82.75 $\pm$ 4.32} \\ \cline{2-5}
                                                                                 & 3              & 61.35 $\pm$ 7.21 & 84.59 $\pm$ 7.65          & \textbf{86.22 $\pm$ 7.15} \\ \hline
      \multirow{3}{*}{Wisconsin}                                                 & 1              & 51.35 $\pm$ 5.05 & \textbf{83.08 $\pm$ 3.49} & 82.31 $\pm$ 2.84          \\ \cline{2-5}
                                                                                 & 2              & 53.85 $\pm$ 3.51 & 85.77 $\pm$ 4.37          & \textbf{86.35 $\pm$ 3.45} \\ \cline{2-5}
                                                                                 & 3              & 59.02 $\pm$ 7.81 & \textbf{88.24 $\pm$ 2.45} & 87.25 $\pm$ 2.12          \\ \hline
    \end{tabular}
  \end{adjustbox}
  \label{tab:node-classification-results}
  \vspace{-0.6cm}
\end{table}
\begin{figure}[t]
  \centering
  \captionsetup[subfigure]{font=scriptsize}
  \vspace{-0.3cm}
  \subfloat[Original node features]{\includegraphics[width = 0.24\textwidth]{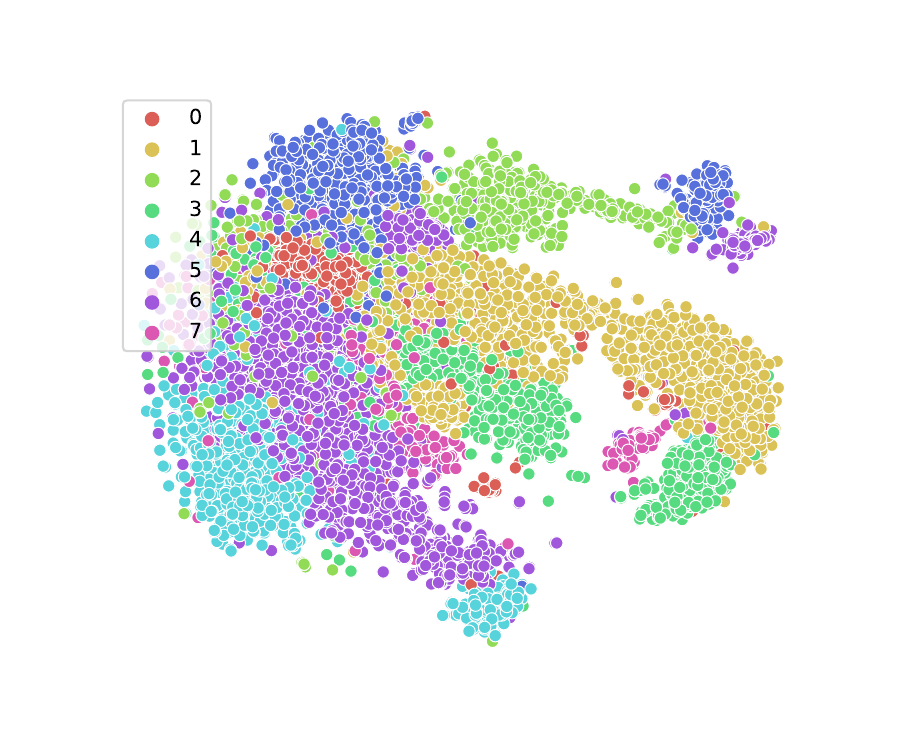}}
  \hfill
  \subfloat[Node embeddings]{\includegraphics[width = 0.24\textwidth]{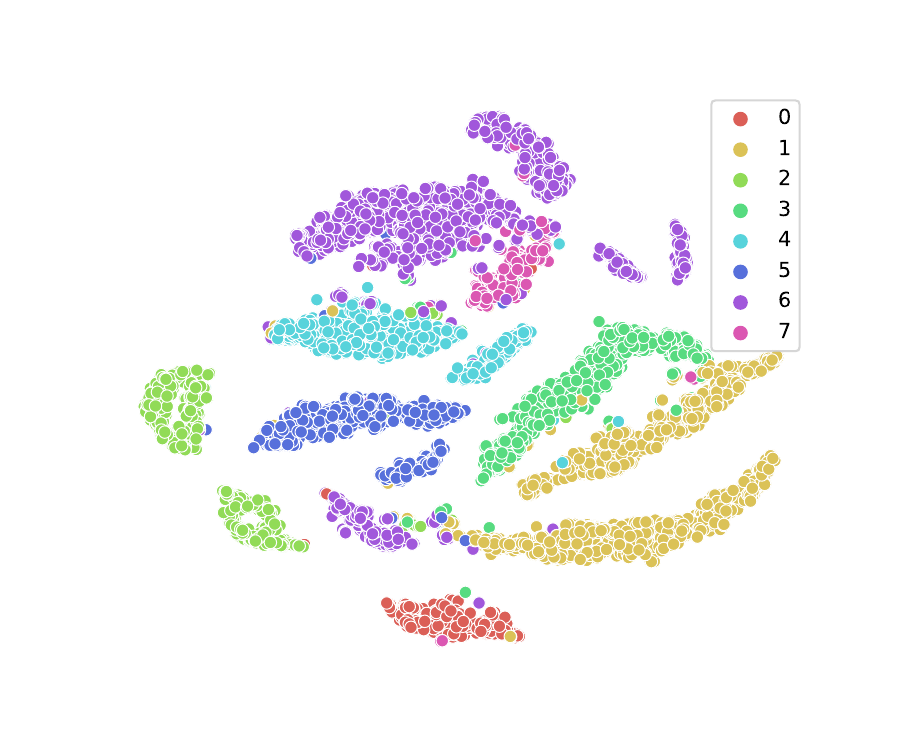}}
  \vspace{-0.1cm}
  \caption{T-SNE visualization of Amazon Photo.}
  \label{fig:tsne}
  \vspace{-0.6cm}
\end{figure}

\subsubsection{Models and Settings}
For node classification tasks, we use the same training setting with the link prediction task. We perform full-supervised transductive learning, comparing three often used split ways\cite{he2021bernnet,song2023ordered,luan2022revisiting} as \cite{pitfall} showing different splits strategies greatly affects results, where use the same split ratios but split each class's nodes 6:2:2, only split training set with balance numbers of each class and randomly split the nodes into 6:2:2. We denote them as split 1,2 and 3. Moreover, for the supervised information can be used in training set, one post process can be performed. Therefore, we also evaluate the results of drop negative edge which connect two nodes with same label in training set in Table~\ref{tab:node-classification-results}, denoted as SGCN$+$. The ratio of wrong edges are shown in Figure~\ref{fig:WrongInTrain} and Figure~\ref{fig:WrongInTrainHete} as one reflection of the quality of negative edges.

\subsubsection{Results for Node Classification}
As results shown in Table~\ref{tab:node-classification-results}, our method significantly enhances the performance of the baseline models. It demonstrates that our method is effective in node classification task. One visualization example of Amazon photo with T-SNE\cite{tsne} is shown in Figure~\ref{fig:tsne}. We can see that our method can help GNN learn more accurate and reliable node representations.

We furthermore compare our results and current state-of-art methods, showing in Table~\ref{tab:sota}. In Appendix C, we list their metrics. In most homophilic dataset, we achieve better results than state-of-the-art method. In heterophilic dataset, we significantly shorten the gap between state-of-the-art method and GCN, even through GCN is not designed for heterophilic graph. It demonstrates that our method is effectively boosting baseline model in both homophilic and heterophilic graph.
\begin{table}[h]
  \centering
  \vspace{-0.5cm}
  \caption{Comparing how much our method shorten the gap between GCN and state-of-the-art method}
  \vspace{-0.1cm}
  \begin{adjustbox}{width=\columnwidth,center}
    \begin{tabular}{|c|c|c|c|c|}
      \hline
      Dataset                                                    & Reference                 & GCN      & Ours     & Improvent \\ \hline
      \begin{tabular}[c]{@{}c@{}}Amazon\\ Photo\end{tabular}     & \cite{he2021bernnet}      & -1.02\%  & 0.96\%   & 1.98\%    \\ \hline
      \begin{tabular}[c]{@{}c@{}}Amazon\\ Computers\end{tabular} & \cite{math10081262}       & -0.43\%  & 0.10\%   & 0.53\%    \\ \hline
      \begin{tabular}[c]{@{}c@{}}Coauthor\\ CS\end{tabular}      & \cite{10251551}      & -0.27\%  & 1.36\%   & 1.62\%    \\ \hline
      Actor                                                      & \cite{9645300}            & -21.37\% & -3.77\%  & 17.60\%   \\ \hline
      Chameleon                                                  & \cite{luan2022revisiting} & -47.28\% & -31.30\% & 15.98\%   \\ \hline
      Cornell                                                    & \cite{luan2022revisiting} & -46.53\% & -20.05\% & 26.48\%   \\ \hline
      Texas                                                      & \cite{luan2022revisiting} & -38.12\% & -14.30\% & 23.82\%   \\ \hline
      Wisconsion                                                 & \cite{luan2022revisiting} & -44.77\% & -11.44\% & 33.33\%   \\ \hline
    \end{tabular}
  \end{adjustbox}
  \label{tab:sota}
  \vspace{-0.7cm}
\end{table}

\subsection{Generalization Ability Discussion}
To further understand the improving by negative edges, we'd like to discuss the generalization ability it brings. Negative edges can bridge training and testing, greatly benefiting the transductive learning. The ratios of negative edges which connect nodes in training and rest sets are shown in Figure~\ref{fig:bridge} and Figure~\ref{fig:bridgeHete}. We can see that the negative edge bridge is effective in all datasets. It demonstrates that our method help GNN gain more information from negative edges, which is helpful for generalization ability.

\section{Conclusion}\
We have performed several empirical evaluations which shown the effectiveness of negative pseudo partial labels and the improving on GNN models via message-passing with negative edges. The negative pseudo partial labels extraction module are consisted by simple and effective parts.

Our method is suitable for attribute graph with different class of nodes. High accurate and reliable negative pseudo partial labels for node pairs, which is the key to our method, can form negative edges for message-passing. The negative edges can help GNN models to learn more accurate and reliable node representations, enhancing generalization ability by aggregating long and abundant information. Empirical evaluations on several datasets for both link prediction and node classification tasks have shown the effectiveness of our method. Discussion on heterophilic datasets is also provided, which shows that our method also have ability boosting GNN models on heterophilic datasets.

\bibliographystyle{IEEEtran}
\bibliography{ref}
\newpage
\onecolumn
\appendix
\section*{A:Signed Graph Neural Neworks}
\label{appendix:sgcn}
SGCN (Signed Graph Neural Network) is a typical GNN for signed graph, which can be formulated as below:
\begin{equation}
    \begin{split}
        \mathbf{Z}_v^{(pos,k)} &= \sigma(\mathbf{W}^{(pos)} \lbrack \sum_{u\in \mathcal{N^+}(v)}\frac{\mathbf{Z}_u^{(pos,k-1)}}{|\mathcal{N^+}(v)|},\sum_{j\in \mathcal{N^-}(v)}\frac{\mathbf{Z}_k^{(n,k-1)} }{|\mathcal{N^-}(v)|},\mathbf{Z}_v^{(pos,k-1)}\rbrack),\\
        \mathbf{Z}_v^{(neg,k)} &= \sigma(\mathbf{W}^{(neg)} \lbrack \sum_{u\in \mathcal{N^+}(v)}\frac{\mathbf{Z}_u^{(neg,k-1)}}{|\mathcal{N^+}(v)|},\sum_{j\in \mathcal{N^-}(v)}\frac{\mathbf{Z}_k^{(pos,k-1)} }{|\mathcal{N^-}(v)|},\mathbf{Z}_v^{(neg,k-1)}\rbrack),\\
        \mathbf{Z}_v^{(k)} &= \lbrack \mathbf{Z}_v^{(pos,k-1)},\mathbf{Z}_v^{(neg,k-1)} \rbrack,
    \end{split}
\end{equation}
where $\mathcal{N^+}(v)$ is the neighborhoods connected by positive edges, $\mathcal{N^-}(v)$ is the neighborhoods connected by negative edges.
\section*{B: Datasets}
\label{appendix:datasets}
Amazon Photo and Amazon Computer are two subsets of Amazon product co-purchasing network. The nodes represent products and edges represent the two products are co-purchased frequently. The features of nodes are extracted from the product reviews as bag-of-words vectors. The classes of nodes are the categories of products.

CS is a subset of the Microsoft Academic Graph. The nodes represent authors and edges represent the co-authorship relationship between authors in the same paper. Node features represent paper keywords for each author’s papers, and class labels indicate most active fields of study for each author.

Actor, co-occurrence network, is the actor-only induced subgraph of the film-director-actor-writer network. The nodes represent actors and edges represent the co-occurrence relationship between actors in the Wikipedia pages. The features of nodes are extracted from the Wikipedia pages keywords. The classes of nodes are the categories of actors.

Cora and CiteSeer are two classical citation network datasets. Nodes represent documents and edges represent citation links.

Both Chameleon and Squirrel datasets are from the same source. The nodes represent the Wikipedia pages and edges represent the hyperlinks between pages. The features of nodes are extracted from the Wikipedia pages keywords.

Cornell, Texas, and Wisconsin are three subsets of the webpage dataset collected by computer science departments of various universities by Carnegie Mellon University. The nodes represent webpages and edges represent the hyperlinks between webpages. The features of nodes are bag-of-words vectors extracted from the webpages. The classes of nodes are the categories of webpages, student, project, course, staff, and faculty.
\section*{C: State-of-the-art Results Comparing}
\label{appendix:sota}
In Table\ref{app:sotatable}, we list the metrics of GCN, our methods and the state-of-the art methods.
\begin{table}[h]
    \centering
    \caption{Comparing the metrics of GCN, our methods and state-of-the-art method}
    \begin{tabular}{|c|c|c|c|c|c|c|c|}
    \hline
    Dataset   & GCN   & SGCN  & SGCN+          & State-of-the-art Methods           & Baseline Gap & Our Gap & Improve \\ \hline
    Photo     & 93.88 & 95.66 & \textbf{95.76} & 94.85          & -1.02\%       & 0.96\%   & 1.98\%  \\ \hline
    Computers & 90.35 & 90.82 & \textbf{90.83} & 90.74          & -0.43\%       & 0.10\%   & 0.53\%  \\ \hline
    CS        & 93.45 & 94.81 & \textbf{94.97} & 93.70          & -0.27\%       & 1.36\%   & 1.62\%  \\ \hline
    Actor     & 29.80 & 36.47 & 36.32          & \textbf{37.90} & -21.37\%      & -3.77\%  & 17.60\% \\ \hline
    Chameleon & 40.11 & 52.27 & 51.4           & \textbf{76.08} & -47.28\%      & -31.30\% & 15.98\% \\ \hline
    Cornell   & 51.28 & 76.15 & 76.67          & \textbf{95.90} & -46.53\%      & -20.05\% & 26.48\% \\ \hline
    Texas     & 59.75 & 82.75 & 82.75          & \textbf{96.56} & -38.12\%      & -14.30\% & 23.82\% \\ \hline
    Wisconsin & 53.85 & 85.77 & 86.35          & \textbf{97.50} & -44.77\%      & -11.44\% & 33.33\% \\ \hline
    \end{tabular}
    \label{app:sotatable}
    \end{table}
\section*{D: Experiment Platform}
We conduct our experiments on a server with NVIDIA Tesla V100 (16G) GPU with CUDA version of 11.7 and Intel(R) Xeon(R) Gold 6151 CPU @ 3.00GHz. The operating system is Ubuntu 20.04.5 LTS. The programming language is Python 3.10.10. The deep learning framework is PyTorch 2.0.0. The graph neural network library is PyG 2.3.0.

\section*{E: Final Hyperparameters}
\subsection*{Link Prediction}
The final hyperparameters for models using GCN as encoder are listed in Table~\ref{link}, ~\ref{link2}. Final model are selected based on metrics on valid set with the best performance.

\begin{table}[h]
    \centering
    \caption{Final hyperparameters for each dataset and GCN model on Link Prediction and Node Classification tasks}
    \begin{tabular}{|c|c|c|c|c|}
        \hline
        \multirow{2}{*}{Dataset}                                             &
        \multirow{2}{*}{Model}                                               &
        \multirow{2}{*}{Partial Label Number}                                &
        \multirow{2}{*}{Learning  Rate}                                      &
        \multirow{2}{*}{Weight decay}                                          \\
                                                                             &
                                                                             &
                                                                             &
                                                                             &
        \\ \hline
        Photo                                                                &
        \multirow{7}{*}{\begin{tabular}[c]{@{}c@{}}GAE\\ GCN\end{tabular}}   &
        /                                                                    &
        0.01                                                                 &
        5e-1                                                                   \\ \cline{1-1} \cline{3-5}
        Computers                                                            &
                                                                             &
        /                                                                    &
        0.01                                                                 &
        5e-1                                                                   \\ \cline{1-1} \cline{3-5}
        CS                                                                   &
                                                                             &
        /                                                                    &
        0.01                                                                 &
        1e-4                                                                   \\ \cline{1-1} \cline{3-5}
        Cora                                                                 &
                                                                             &
        /                                                                    &
        0.01                                                                 &
        5e-1                                                                   \\ \cline{1-1} \cline{3-5}
        CiteSeer                                                             &
                                                                             &
        /                                                                    &
        0.01                                                                 &
        1e-3                                                                   \\ \cline{1-1} \cline{3-5}
        Chameleon                                                            &
                                                                             &
        /                                                                    &
        0.01                                                                 &
        5e-1                                                                   \\ \cline{1-1} \cline{3-5}
        Squirrel                                                             &
                                                                             &
        /                                                                    &
        0.01                                                                 &
        5e-1                                                                   \\ \hline
        \multirow{2}{*}{Dataset}                                             &
        \multirow{2}{*}{Model}                                               &
        \multirow{2}{*}{Partial Label Number}                                &
        \multirow{2}{*}{Learning  Rate}                                      &
        \multirow{2}{*}{Weight decay}                                          \\
                                                                             &
                                                                             &
                                                                             &
                                                                             &
        \\ \hline
        Photo                                                                &
        \multirow{7}{*}{\begin{tabular}[c]{@{}c@{}}GAE\\ SGCN\end{tabular}}  &
        4                                                                    &
        0.01                                                                 &
        5e-1                                                                   \\ \cline{1-1} \cline{3-5}
        Computers                                                            &
                                                                             &
        5                                                                    &
        0.01                                                                 &
        5e-1                                                                   \\ \cline{1-1} \cline{3-5}
        CS                                                                   &
                                                                             &
        7                                                                    &
        0.01                                                                 &
        1e-3                                                                   \\ \cline{1-1} \cline{3-5}
        Cora                                                                 &
                                                                             &
        3                                                                    &
        0.01                                                                 &
        5e-1                                                                   \\ \cline{1-1} \cline{3-5}
        CiteSeer                                                             &
                                                                             &
        2                                                                    &
        0.01                                                                 &
        1e-3                                                                   \\ \cline{1-1} \cline{3-5}
        Chameleon                                                            &
                                                                             &
        2                                                                    &
        0.01                                                                 &
        5e-1                                                                   \\ \cline{1-1} \cline{3-5}
        Squirrel                                                             &
                                                                             &
        2                                                                    &
        0.01                                                                 &
        5e-1                                                                   \\ \hline
        \multirow{2}{*}{Dataset}                                             &
        \multirow{2}{*}{Model}                                               &
        \multirow{2}{*}{Partial Label Number}                                &
        \multirow{2}{*}{Learning  Rate}                                      &
        \multirow{2}{*}{Weight decay}                                          \\
                                                                             &
                                                                             &
                                                                             &
                                                                             &
        \\ \hline
        Photo                                                                &
        \multirow{7}{*}{\begin{tabular}[c]{@{}c@{}}VGAE\\ GCN\end{tabular}}  &
        /                                                                    &
        0.01                                                                 &
        1e-4                                                                   \\ \cline{1-1} \cline{3-5}
        Computers                                                            &
                                                                             &
        /                                                                    &
        0.01                                                                 &
        5e-1                                                                   \\ \cline{1-1} \cline{3-5}
        CS                                                                   &
                                                                             &
        /                                                                    &
        0.01                                                                 &
        1e-4                                                                   \\ \cline{1-1} \cline{3-5}
        Cora                                                                 &
                                                                             &
        /                                                                    &
        0.01                                                                 &
        5e-1                                                                   \\ \cline{1-1} \cline{3-5}
        CiteSeer                                                             &
                                                                             &
        /                                                                    &
        0.01                                                                 &
        1e-4                                                                   \\ \cline{1-1} \cline{3-5}
        Chameleon                                                            &
                                                                             &
        /                                                                    &
        0.01                                                                 &
        5e-1                                                                   \\ \cline{1-1} \cline{3-5}
        Squirrel                                                             &
                                                                             &
        /                                                                    &
        0.01                                                                 &
        5e-1                                                                   \\ \hline
        \multirow{2}{*}{Dataset}                                             &
        \multirow{2}{*}{Model}                                               &
        \multirow{2}{*}{Partial Label Number}                                &
        \multirow{2}{*}{Learning  Rate}                                      &
        \multirow{2}{*}{Weight decay}                                          \\
                                                                             &
                                                                             &
                                                                             &
                                                                             &
        \\ \hline
        Photo                                                                &
        \multirow{7}{*}{\begin{tabular}[c]{@{}c@{}}VGAE\\ SGCN\end{tabular}} &
        4                                                                    &
        0.01                                                                 &
        1e-4                                                                   \\ \cline{1-1} \cline{3-5}
        Computers                                                            &
                                                                             &
        5                                                                    &
        0.01                                                                 &
        5e-1                                                                   \\ \cline{1-1} \cline{3-5}
        CS                                                                   &
                                                                             &
        7                                                                    &
        0.01                                                                 &
        5e-1                                                                   \\ \cline{1-1} \cline{3-5}
        Cora                                                                 &
                                                                             &
        3                                                                    &
        0.01                                                                 &
        5e-1                                                                   \\ \cline{1-1} \cline{3-5}
        CiteSeer                                                             &
                                                                             &
        2                                                                    &
        0.01                                                                 &
        1e-4                                                                   \\ \cline{1-1} \cline{3-5}
        Chameleon                                                            &
                                                                             &
        2                                                                    &
        0.01                                                                 &
        5e-1                                                                   \\ \cline{1-1} \cline{3-5}
        Squirrel                                                             &
                                                                             &
        2                                                                    &
        0.01                                                                 &
        5e-1                                                                   \\ \hline
    \end{tabular}
    \label{link}
\end{table}
% \usepackage{multirow}
\begin{table}[]
    \centering
    \caption{Final hyperparameters for each dataset and GNCN model on Link Prediction tasks}
    \begin{tabular}{|c|c|c|c|c|}
        \hline
        \multirow{2}{*}{Dataset}                                              &
        \multirow{2}{*}{Model}                                                &
        \multirow{2}{*}{Partial Label Number}                                 &
        \multirow{2}{*}{Learning Rate}                                        &
        \multirow{2}{*}{Weight Decay}                                           \\
                                                                              &
                                                                              &
                                                                              &
                                                                              &
        \\ \hline
        Photo                                                                 &
        \multirow{7}{*}{\begin{tabular}[c]{@{}c@{}}GAE\\ GNCN\end{tabular}}   &
        /                                                                     &
        0.1                                                                   &
        1e-3                                                                    \\ \cline{1-1} \cline{3-5}
        Computers                                                             &
                                                                              &
        /                                                                     &
        0.1                                                                   &
        1e-3                                                                    \\ \cline{1-1} \cline{3-5}
        CS                                                                    &
                                                                              &
        /                                                                     &
        0.1                                                                   &
        1e-2                                                                    \\ \cline{1-1} \cline{3-5}
        Cora                                                                  &
                                                                              &
        /                                                                     &
        0.1                                                                   &
        1e-3                                                                    \\ \cline{1-1} \cline{3-5}
        CiteSeer                                                              &
                                                                              &
        /                                                                     &
        0.1                                                                   &
        1e-3                                                                    \\ \cline{1-1} \cline{3-5}
        Chameleon                                                             &
                                                                              &
        /                                                                     &
        0.1                                                                   &
        5e-1                                                                    \\ \cline{1-1} \cline{3-5}
        Squirrel                                                              &
                                                                              &
        /                                                                     &
        0.1                                                                   &
        1e-3                                                                    \\ \hline
        \multirow{2}{*}{Dataset}                                              &
        \multirow{2}{*}{Model}                                                &
        \multirow{2}{*}{Partial Label Number}                                 &
        \multirow{2}{*}{Learning Rate}                                        &
        \multirow{2}{*}{Weight Decay}                                           \\
                                                                              &
                                                                              &
                                                                              &
                                                                              &
        \\ \hline
        Photo                                                                 &
        \multirow{7}{*}{\begin{tabular}[c]{@{}c@{}}GAE\\ SGNCN\end{tabular}}  &
        4                                                                     &
        0.1                                                                   &
        1e-3                                                                    \\ \cline{1-1} \cline{3-5}
        Computers                                                             &
                                                                              &
        5                                                                     &
        0.1                                                                   &
        1e-3                                                                    \\ \cline{1-1} \cline{3-5}
        CS                                                                    &
                                                                              &
        7                                                                     &
        0.1                                                                   &
        1e-2                                                                    \\ \cline{1-1} \cline{3-5}
        Cora                                                                  &
                                                                              &
        3                                                                     &
        0.1                                                                   &
        5e-1                                                                    \\ \cline{1-1} \cline{3-5}
        CiteSeer                                                              &
                                                                              &
        2                                                                     &
        0.1                                                                   &
        1e-3                                                                    \\ \cline{1-1} \cline{3-5}
        Chameleon                                                             &
                                                                              &
        2                                                                     &
        0.1                                                                   &
        5e-1                                                                    \\ \cline{1-1} \cline{3-5}
        Squirrel                                                              &
                                                                              &
        2                                                                     &
        0.1                                                                   &
        5e-1                                                                    \\ \hline
        \multirow{2}{*}{Dataset}                                              &
        \multirow{2}{*}{Model}                                                &
        \multirow{2}{*}{Partial Label Number}                                 &
        \multirow{2}{*}{Learning Rate}                                        &
        \multirow{2}{*}{Weight Decay}                                           \\
                                                                              &
                                                                              &
                                                                              &
                                                                              &
        \\ \hline
        Photo                                                                 &
        \multirow{7}{*}{\begin{tabular}[c]{@{}c@{}}VGAE\\ GNCN\end{tabular}}  &
        /                                                                     &
        0.1                                                                   &
        1e-3                                                                    \\ \cline{1-1} \cline{3-5}
        Computers                                                             &
                                                                              &
        /                                                                     &
        0.1                                                                   &
        1e-4                                                                    \\ \cline{1-1} \cline{3-5}
        CS                                                                    &
                                                                              &
        /                                                                     &
        0.1                                                                   &
        1e-4                                                                    \\ \cline{1-1} \cline{3-5}
        Cora                                                                  &
                                                                              &
        /                                                                     &
        0.1                                                                   &
        5e-1                                                                    \\ \cline{1-1} \cline{3-5}
        CiteSeer                                                              &
                                                                              &
        /                                                                     &
        0.1                                                                   &
        1e-3                                                                    \\ \cline{1-1} \cline{3-5}
        Chameleon                                                             &
                                                                              &
        /                                                                     &
        0.1                                                                   &
        1e-4                                                                    \\ \cline{1-1} \cline{3-5}
        Squirrel                                                              &
                                                                              &
        /                                                                     &
        0.1                                                                   &
        1e-4                                                                    \\ \hline
        \multirow{2}{*}{Dataset}                                              &
        \multirow{2}{*}{Model}                                                &
        \multirow{2}{*}{Partial Label Number}                                 &
        \multirow{2}{*}{Learning Rate}                                        &
        \multirow{2}{*}{Weight Decay}                                           \\
                                                                              &
                                                                              &
                                                                              &
                                                                              &
        \\ \hline
        Photo                                                                 &
        \multirow{7}{*}{\begin{tabular}[c]{@{}c@{}}VGAE\\ SGNCN\end{tabular}} &
        4                                                                     &
        0.1                                                                   &
        1e-3                                                                    \\ \cline{1-1} \cline{3-5}
        Computers                                                             &
                                                                              &
        5                                                                     &
        0.1                                                                   &
        1e-4                                                                    \\ \cline{1-1} \cline{3-5}
        CS                                                                    &
                                                                              &
        /                                                                     &
        /                                                                     &
        /                                                                       \\ \cline{1-1} \cline{3-5}
        Cora                                                                  &
                                                                              &
        3                                                                     &
        0.1                                                                   &
        1e-4                                                                    \\ \cline{1-1} \cline{3-5}
        CiteSeer                                                              &
                                                                              &
        2                                                                     &
        0.1                                                                   &
        1e-3                                                                    \\ \cline{1-1} \cline{3-5}
        Chameleon                                                             &
                                                                              &
        2                                                                     &
        0.1                                                                   &
        1e-4                                                                    \\ \cline{1-1} \cline{3-5}
        Squirrel                                                              &
                                                                              &
        2                                                                     &
        0.1                                                                   &
        5e-1                                                                    \\ \hline
    \end{tabular}
    \label{link2}
\end{table}
\subsection*{Node Classification}
For node classification, 10 times experiments are conducted form each dataset and split way. The final hyperparameters are shown in Table \ref{node}.
% \usepackage{multirow}
\begin{table}[]
    \centering
    \caption{Final hyperparameters for each dataset on Node Classification tasks}
    \begin{tabular}{|c|c|c|c|c|c|}
        \hline
        Dataset                    & Split & Model                  & Partial Label Number & Learning Rate & Weight Decay \\ \hline
        \multirow{3}{*}{Photo}     & 1     & \multirow{27}{*}{GCN}  & /                    & 0.01          & 0e+0         \\ \cline{2-2} \cline{4-6}
                                   & 2     &                        & /                    & 0.001         & 0e+0         \\ \cline{2-2} \cline{4-6}
                                   & 3     &                        & /                    & 0.001         & 0e+0         \\ \cline{1-2} \cline{4-6}
        \multirow{3}{*}{Computers} & 1     &                        & /                    & 0.001         & 0e+0         \\ \cline{2-2} \cline{4-6}
                                   & 2     &                        & /                    & 0.001         & 0e+0         \\ \cline{2-2} \cline{4-6}
                                   & 3     &                        & /                    & 0.001         & 0e+0         \\ \cline{1-2} \cline{4-6}
        \multirow{3}{*}{CS}        & 1     &                        & /                    & 0.001         & 1e-3         \\ \cline{2-2} \cline{4-6}
                                   & 2     &                        & /                    & 0.01          & 1e-3         \\ \cline{2-2} \cline{4-6}
                                   & 3     &                        & /                    & 0.01          & 1e-3         \\ \cline{1-2} \cline{4-6}
        \multirow{3}{*}{Actor}     & 1     &                        & /                    & 0.001         & 1e-2         \\ \cline{2-2} \cline{4-6}
                                   & 2     &                        & /                    & 0.001         & 1e-2         \\ \cline{2-2} \cline{4-6}
                                   & 3     &                        & /                    & 0.01          & 1e-2         \\ \cline{1-2} \cline{4-6}
        \multirow{3}{*}{Chameleon} & 1     &                        & /                    & 0.05          & 0e+0         \\ \cline{2-2} \cline{4-6}
                                   & 2     &                        & /                    & 0.001         & 5e-2         \\ \cline{2-2} \cline{4-6}
                                   & 3     &                        & /                    & 0.05          & 0e+0         \\ \cline{1-2} \cline{4-6}
        \multirow{3}{*}{Squirrel}  & 1     &                        & /                    & 0.01          & 1e-3         \\ \cline{2-2} \cline{4-6}
                                   & 2     &                        & /                    & 0.001         & 1e-3         \\ \cline{2-2} \cline{4-6}
                                   & 3     &                        & /                    & 0.001         & 1e-3         \\ \cline{1-2} \cline{4-6}
        \multirow{3}{*}{Cornell}   & 1     &                        & /                    & 0.01          & 5e-2         \\ \cline{2-2} \cline{4-6}
                                   & 2     &                        & /                    & 0.05          & 5e-2         \\ \cline{2-2} \cline{4-6}
                                   & 3     &                        & /                    & 0.01          & 1e-2         \\ \cline{1-2} \cline{4-6}
        \multirow{3}{*}{Texas}     & 1     &                        & /                    & 0.05          & 5e-2         \\ \cline{2-2} \cline{4-6}
                                   & 2     &                        & /                    & 0.01          & 5e-2         \\ \cline{2-2} \cline{4-6}
                                   & 3     &                        & /                    & 0.05          & 5e-2         \\ \cline{1-2} \cline{4-6}
        \multirow{3}{*}{Wisconsin} & 1     &                        & /                    & 0.05          & 5e-2         \\ \cline{2-2} \cline{4-6}
                                   & 2     &                        & /                    & 0.01          & 5e-2         \\ \cline{2-2} \cline{4-6}
                                   & 3     &                        & /                    & 0.001         & 5e-2         \\ \hline
        Dataset                    & Split & Model                  & Partial Label Number & Learning Rate & Weight Decay \\ \hline
        \multirow{3}{*}{Photo}     & 1     & \multirow{27}{*}{SGCN} & 4                    & 0.001         & 1e-2         \\ \cline{2-2} \cline{4-6}
                                   & 2     &                        & 4                    & 0.001         & 1e-2         \\ \cline{2-2} \cline{4-6}
                                   & 3     &                        & 4                    & 0.001         & 1e-2         \\ \cline{1-2} \cline{4-6}
        \multirow{3}{*}{Computers} & 1     &                        & 5                    & 0.001         & 1e-2         \\ \cline{2-2} \cline{4-6}
                                   & 2     &                        & 5                    & 0.001         & 1e-2         \\ \cline{2-2} \cline{4-6}
                                   & 3     &                        & 5                    & 0.001         & 1e-2         \\ \cline{1-2} \cline{4-6}
        \multirow{3}{*}{CS}        & 1     &                        & 6                    & 0.001         & 1e-3         \\ \cline{2-2} \cline{4-6}
                                   & 2     &                        & 6                    & 0.001         & 1e-3         \\ \cline{2-2} \cline{4-6}
                                   & 3     &                        & 6                    & 0.01          & 1e-3         \\ \cline{1-2} \cline{4-6}
        \multirow{3}{*}{Actor}     & 1     &                        & 2                    & 0.01          & 1e-2         \\ \cline{2-2} \cline{4-6}
                                   & 2     &                        & 2                    & 0.001         & 1e-3         \\ \cline{2-2} \cline{4-6}
                                   & 3     &                        & 2                    & 0.1           & 1e-2         \\ \cline{1-2} \cline{4-6}
        \multirow{3}{*}{Chameleon} & 1     &                        & 2                    & 0.001         & 0e+0         \\ \cline{2-2} \cline{4-6}
                                   & 2     &                        & 2                    & 0.001         & 0e+0         \\ \cline{2-2} \cline{4-6}
                                   & 3     &                        & 2                    & 0.001         & 0e+0         \\ \cline{1-2} \cline{4-6}
        \multirow{3}{*}{Squirrel}  & 1     &                        & 2                    & 0.01          & 1e-3         \\ \cline{2-2} \cline{4-6}
                                   & 2     &                        & 2                    & 0.01          & 1e-3         \\ \cline{2-2} \cline{4-6}
                                   & 3     &                        & 2                    & 0.01          & 1e-3         \\ \cline{1-2} \cline{4-6}
        \multirow{3}{*}{Cornell}   & 1     &                        & 2                    & 0.05          & 5e-2         \\ \cline{2-2} \cline{4-6}
                                   & 2     &                        & 2                    & 0.1           & 1e-2         \\ \cline{2-2} \cline{4-6}
                                   & 3     &                        & 2                    & 0.01          & 1e-2         \\ \cline{1-2} \cline{4-6}
        \multirow{3}{*}{Texas}     & 1     &                        & 2                    & 0.01          & 1e-2         \\ \cline{2-2} \cline{4-6}
                                   & 2     &                        & 2                    & 0.01          & 1e-2         \\ \cline{2-2} \cline{4-6}
                                   & 3     &                        & 2                    & 0.01          & 1e-3         \\ \cline{1-2} \cline{4-6}
        \multirow{3}{*}{Wisconsin} & 1     &                        & 2                    & 0.01          & 1e-3         \\ \cline{2-2} \cline{4-6}
                                   & 2     &                        & 2                    & 0.01          & 5e-2         \\ \cline{2-2} \cline{4-6}
                                   & 3     &                        & 2                    & 0.01          & 5e-2         \\ \hline
    \end{tabular}
    \label{node}
\end{table}

\vfill

\end{document}